\def\year{2020}\relax
\documentclass[letterpaper]{article} 
\usepackage{aaai20}  
\usepackage{times}  
\usepackage{helvet} 
\usepackage{courier}  
\usepackage[hyphens]{url}  
\usepackage{graphicx} 
\usepackage{graphicx}  
\usepackage{amsmath}
\usepackage{amsthm}
\usepackage{amsfonts}
\usepackage{algorithm}
\usepackage{algorithmic}
\usepackage{subfigure}
\usepackage{epstopdf}
\usepackage{enumitem}
\usepackage{multirow}
\usepackage{diagbox}

\newtheorem{lemma}{Lemma}
\newtheorem{theorem}{Theorem}
\newtheorem{proposition}[theorem]{Proposition}

\def\X{\mathbf{X}}

\title{Improved Subsampled Randomized Hadamard Transform for Linear SVM}
\author{
   Zijian Lei\\
   Department of Computer Science, \\
   Hong Kong Baptist University, \\
   Hong Kong SAR, China\\
   cszjlei@comp.hkbu.edu.hk
   \And
   Liang Lan\\
   Department of Computer Science, \\
   Hong Kong Baptist University, \\
   Hong Kong SAR, China\\
   lanliang@comp.hkbu.edu.hk
}

\begin{document}
\maketitle
\begin{abstract}
Subsampled Randomized Hadamard Transform (SRHT), a popular random projection method that can efficiently project a $d$-dimensional data into $r$-dimensional space ($r \ll d$) in $O(dlog(d))$ time, has been widely used to address the challenge of high-dimensionality in machine learning. SRHT works by rotating the input data matrix $\mathbf{X} \in \mathbb{R}^{n \times d}$ by Randomized Walsh-Hadamard Transform followed with a subsequent uniform column sampling on the rotated matrix. Despite the advantages of SRHT, one limitation of SRHT is that it generates the new low-dimensional embedding without considering any specific properties of a given dataset. Therefore, this data-independent random projection method may result in inferior and unstable performance when used for a particular machine learning task, e.g., classification. To overcome this limitation, we analyze the effect of using SRHT for random projection in the context of linear SVM classification. Based on our analysis, we propose importance sampling and deterministic top-$r$ sampling to produce effective low-dimensional embedding instead of uniform sampling SRHT. In addition, we also proposed a new supervised non-uniform sampling method. Our experimental results have demonstrated that our proposed methods can achieve higher classification accuracies than SRHT and other random projection methods on six real-life datasets.
\end{abstract}

\section{Introduction}
Classification is one of the fundamental machine learning tasks. In the era of big data, huge amounts of high-dimensional data become common in a wide variety of applications. As the dimensionality of the data for real-life classification tasks has increased dramatically in recent years, it is essential to develop accurate and efficient classification algorithms for high-dimensional data.

One of the most popular methods to address the high-dimensionality challenge is dimensionality reduction. It allows the classification problems to be efficiently solved in a lower dimensional space. Popular dimensionality reduction methods includes Principal Component Analysis (PCA) \cite{jolliffe2011principal}, Linear Discriminant Analysis (LDA) \cite{mika1999fisher} etc. However, these traditional dimensionality reduction methods are computationally expensive which makes them not suitable for big data. Recently, random projection \cite{bingham2001random} that projects a high-dimensional feature vector $\mathbf{x} \in \mathbb{R}^d$ into a low-dimensional feature $\widetilde{\mathbf{x}} \in \mathbb{R}^r$ ($r \ll d$) using a random orthonormal matrix $\mathbf{R} \in \mathbb{R}^{d \times r}$. Namely, $\widetilde{\mathbf{x}} = \mathbf{R}^{T}\mathbf{x}$. Since random projection methods are (1) computationally efficient while maintaining accurate results; (2) simple to implement in practice; and (3) easy to analyze \cite{mahoney2011randomized,xu2017efficient}, they have attracted significant research attentions in recent years.

The theoretical foundation of random projection is the Johnson-Lindenstrauss Lemma (JL lemma) \cite{johnson1986extensions}. JL lemma proves that in Euclidean space, high-dimensional data can be randomly projected into lower dimensional space while the pairwise distances between all data points are preserved with a high probability. Based on the JL lemma, several different ways are proposed for constructing the random matrix $\mathbf{R}$: (1) Gaussian Matrix \cite{dasgupta1999elementary}: each entry in $\mathbf{R}$ is generated from a Gaussian distribution with mean equals to 0 and variance equals to $\frac{1}{d}$; (2) Achlioptas matrix \cite{achlioptas2003database}: each entry in $\mathbf{R}$ is generated from \{$-1$, 0, 1\} from a discrete distribution. This method generates a sparse random matrix $\mathbf{R}$. Therefore it requires less memory and computation cost; (3) sparse embedding matrix (or called count sketch matrix) \cite{clarkson2017low} and its similar variant feature hashing \cite{weinberger2009feature,freksen2018fully}: for each row $\mathbf{R}$, only one column is randomly selected and assign either 1 or $-1$ with probability 0.5 to this entry; and (4) Subsampled Randomized Hadamard Transform (SRHT) \cite{tropp2011improved}: It uses a highly structured matrix $\mathbf{R}$ for random projection. The procedure of SRHT can be summarized as follows: It first rotates the input matrix $\mathbf{X} \in \mathbb{R}^{n \times d}$ ($n$ is the number of samples, $d$ is the dimensionality of the data) by Randomized Walsh-Hadamard Transform and then uniformly sampling $r$ columns from the rotated matrix. More details of SRHT will be discussed in next section.

Although random projection methods have attracted a great deal of research attention in recent years \cite{sarlos2006improved,choromanski2017unreasonable} and have been studied for regression \cite{lu2013faster}, classification \cite{paul2013random,zhang2013recovering,paul2015feature} and clustering \cite{boutsidis2010random,liu2017sparse}, most of the existing studies focused on data-independent projection. In other words, the random matrix $\mathbf{R}$ was constructed without considering any specific properties of a given dataset. 

In this paper, we focus on SRHT for random projection and study how it affects the classification performance of linear Support Vector Machines (SVM). SRHT achieve dimensionality reduction by uniformly sampling $r$ columns from a rotated version of the input data matrix $\mathbf{X}$. Even though Randomized Walsh-Hadamard transform in SRHT tends to equalize column norms \cite{boutsidis2013improved}, we argue that the importance of different columns in the rotated data matrix are not equal. Therefore, the uniformly random sampling procedure in SRHT would result in low accuracy when used as a dimensionality reduction method in high-dimensional data classification.

To overcome the limitation of SRHT, we propose to produce effective low-dimensional embedding by using non-uniform sampling instead of uniform sampling. To achieve this goal, we first analyze the effect of using SRHT for random projection in linear SVM classification. Based on our analysis, we have proposed importance sampling and deterministic top-$r$ sampling methods to improve SRHT. Furthermore, we also propose a new sampling method by incorporating label information. It samples the columns which can achieve optimal inter-class separability along with the intra-class compactness of the data samples.

Finally, we performed experiments to evaluate our proposed methods on six real-life datasets. Our experimental results clearly demonstrate that our proposed method can obtain higher classification accuracies than SRHT and other three popular random projection methods while only slightly increasing the running time.

\section{Preliminaries}
\subsection{Random Projection}
Given a data matrix $\mathbf{X} \in \mathbb{R}^{n \times d}$, random projection methods reduce the dimensionality of $\mathbf{X}$ by multiplying it by a random orthonormal matrix $\mathbf{R} \in \mathbb{R}^{d \times r}$ with parameter $r \ll d$. The projected data in the low-dimensional space is
\begin{eqnarray}
\mathbf{\widetilde{X}} = \mathbf{XR} \in \mathbb{R}^{n \times r}.
\end{eqnarray}

Note that the matrix $\mathbf{R}$ is randomly generated and is independent of the input data $\mathbf{X}$. The theoretical foundation of random projection is the Johnson-Lindenstrauss Lemma (JL lemma) as shown in following,

\begin{lemma} [Johnson-Lindenstrauss Lemma (JL lemma) \cite{johnson1986extensions}]\label{theorem:JL_lemma}
For any $0<\epsilon<1$ and any integer n, let $r = O(\log n/\epsilon^2)$ and $\mathbf{R} \in \mathbb{R}^{d \times r}$ be a random orthonormal matrix. Then for any set $\mathbf{X}$ of $n$ points in $\mathbb{R}^d$, the following inequality about pairwise distance between any two data points $\mathbf{x}_i$ and $\mathbf{x}_j$ in $\mathbf{X}$ holds true with high probability:
\begin{equation*}
(1-\epsilon)\|\mathbf{x}_i-\mathbf{x}_j\|_2 \le \|\mathbf{R}^{T}\mathbf{x}_i-\mathbf{R}^{T}\mathbf{x}_j\|_2 \le(1+\epsilon)\|\mathbf{x}_i-\mathbf{x}_j\|_2.
\end{equation*}
\end{lemma}
JL lemma proves that in Euclidean space, high-dimensional data can be randomly projected into lower dimensional space while the pairwise distances between all data points are well preserved with high probability. There are several different ways to construct the random projection matrix $\mathbf{R}$. In this paper, we focus on SRHT which uses a structured orthonormal matrix for random projection. The details about SRHT are as follows.

\subsection{Subsampled Randomized Hadamard Transform (SRHT)}
For $d=2^q$where $q$ is any positive integer\footnote{We can ensure this by padding zeros to original data} , SRHT defines a $d \times r$ matrix as:
\begin{equation}\label{eq:SRHT}
\mathbf{R} = \sqrt{\frac{d}{r}}\mathbf{DHS}
\end{equation}
where
\begin{itemize}
    \item $\mathbf{D} \in \mathbb{R}^{d \times d}$ is a diagonal matrix whose elements are independent random signs \{1,$-1$\};
	
	\item $\mathbf{H} \in \mathbb{R}^{d \times d}$ is a normalized Walsh-Hadamard matrix. The Walsh-Hadamard matrix is defined recursively as:  \\
	$   \mathbf{H}_d = \left[\begin{array}{cc}
	\mathbf{H}_{d/2} & \mathbf{H}_{d/2} \\
	\mathbf{H}_{d/2} & -\mathbf{H}_{d/2}
	\end{array}
	\right]$
	with
	$     \mathbf{H}_2 = \left[\begin{array}{cc}
	1 & 1 \\
	1 & -1
	\end{array}
	\right],$ and $\mathbf{H} = \frac{1}{\sqrt{d}}\mathbf{H}_d \in \mathbb{R}^{d\times d}$;
	
    \item $\mathbf{S} \in \mathbb{R}^{d \times r}$ is a subset of randomly sampling $r$ columns from the $d \times d$ identity matrix. The purpose of multiplying $\mathbf{S}$ is to uniformly sample $r$ columns from the rotated data matrix $\mathbf{X}_r = \mathbf{XDH}$.	
\end{itemize}

There are several advantages of using SRHT for random projection. Firstly, we do not need to explicitly represent $\mathbf{H}$ in memory and only need to store a diagonal random sign matrix $\mathbf{D}$ and a sampling matrix $\mathbf{S}$ with $r$ non-zero values. Therefore, the memory cost of storing $\mathbf{R}$ is only $O(d+r)$. Secondly, due to the recursive structure of Hadamard matrix $\mathbf{H}$, matrix multiplication $\mathbf{XR}$ only takes $O(nd\log(d))$ time by using Fast Fourier Transform (FFT) for matrix multiplication \cite{tropp2011improved}. \citeauthor{ailon2009fast} \cite{ailon2009fast} further improve the time complexity of SRHT to $O(nd\log(r))$ if only $r$ columns in $\mathbf{X}_r$ are needed.

\begin{figure*}[bth]
\subfigure[Original data $\mathbf{X}$]{\label{Fig.1.1}\includegraphics[width=0.6\columnwidth]{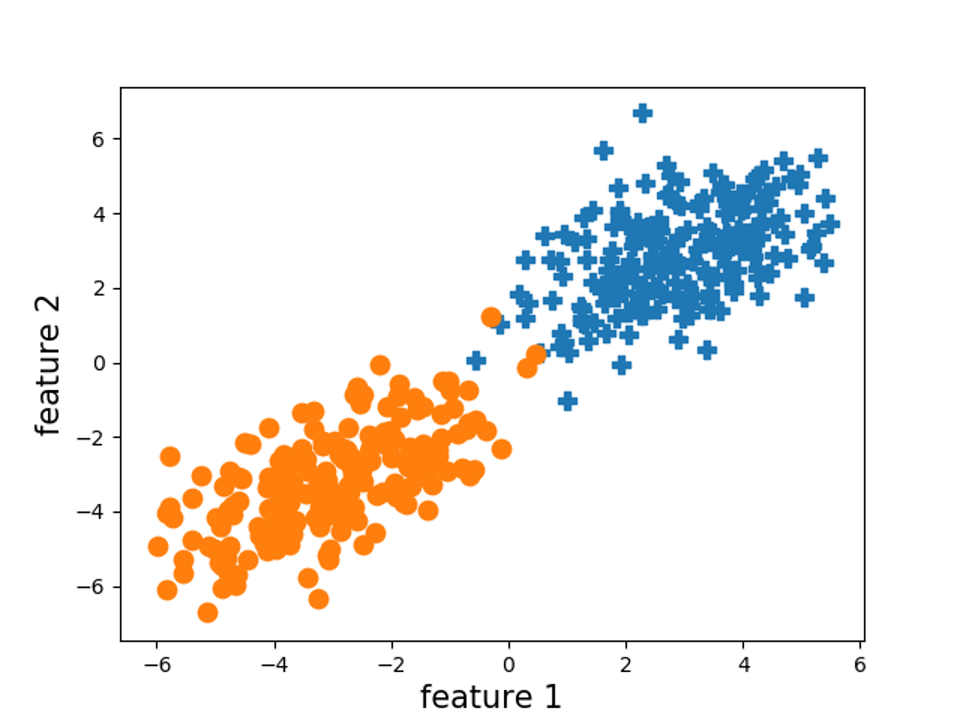}}
\subfigure[Rotated data $\mathbf{X}_r = \mathbf{XDH}$]{\label{Fig.1.2}\includegraphics[width=0.6\columnwidth]{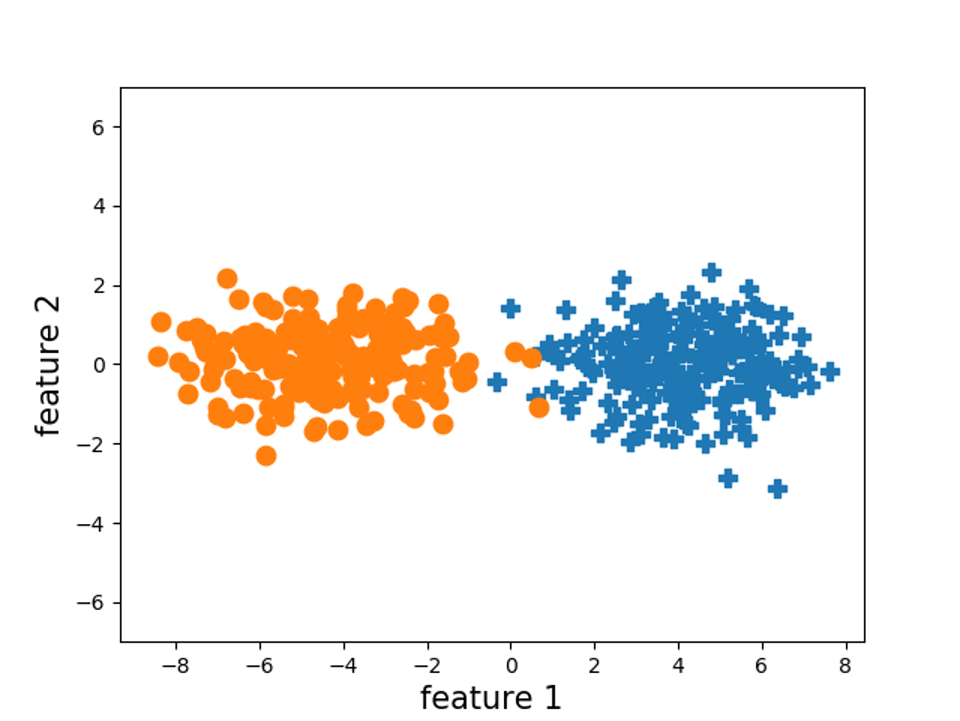}}
\subfigure[Accuracy of SRHT on different runs]{\label{Fig.1.3}\includegraphics[width=0.6\columnwidth]{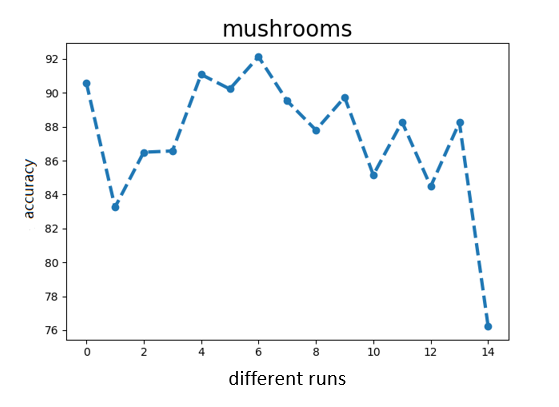}}
\label{Fig.1}
\caption{Limitation of uniform sampling}
\end{figure*}

\section{Methodology}
\subsection{Limitation of Uniform Sampling in SRHT}
Despite the advantages of SRHT as mentioned in the previous section, one limitation of SRHT is that the random matrix $\mathbf{R}$ is constructed without considering any specific properties of input data $\mathbf{X}$. Based on the definition in (\ref{eq:SRHT}), SRHT works by first rotating the input data matrix $\mathbf{X}$ by multiplying it with $\mathbf{DH}$ and then uniformly random sampling $r$ columns from the rotated matrix $\mathbf{X}_r = \mathbf{XDH}$. Since this uniform column sampling method does not consider the underlying data distribution of $\mathbf{X}_r$, it would result in low and unstable accuracy when used as a dimensionality reduction method in a specific classification problem. To illustrate the limitation of uniformly random sampling step in SRHT, we generate a simple two-dimensional synthetic data as shown in Figure $\ref{Fig.1.1}$.
The first class (i.e., class label `\textbf{+}') was generated from $N$($\mu_1 = [3, 3]$, $ \Sigma = \left[\begin{array}{cc}
	1 & 1.75 \\
	1.75 & 1
	\end{array}
	\right]$). The second class (i.e., class label `$\bullet$') was generated from $N$ ($\mu_2 = [-3, -3]$, $ \Sigma = \left[\begin{array}{cc}
	1 & 1.75 \\
	1.75 & 1
	\end{array}
	\right]$). After multiplying with matrix $\mathbf{DH}$, the original data was rotated into new feature space as shown in Figure $\ref{Fig.1.2}$. In this simple illustration example, SRHT might choose the feature that can not distinguish these two classes (i.e, the vertical feature in Figure $\ref{Fig.1.2}$) and then leads to low classification accuracy. In Figure $\ref{Fig.1.3}$, we shown the classification accuracy of using SRHT on a real-life dataset for 15 different runs. It clearly shows that this data-independent random projection method could result in low and unstable performance on this synthetic dataset. In the following sections, we will describe our proposed methods to construct the data-dependent sampling matrix $\mathbf{S}$ in (\ref{eq:SRHT}) by exploiting the underlying data properties from both unsupervised and supervised perspectives.
\subsection{Improved SRHT by unsupervised non-uniform sampling}
First, we present our proposed non-uniform sampling methods from unsupervised perspective. In this paper, we study SRHT in the context of using linear SVM for classification. Assume we are given a training data set $D = \{\mathbf{x}_i, y_i\}_{i=1}^{n}$, where $\mathbf{x}_i$ is a $d$-dimensional input feature vector, $y_i$ is its corresponding class label. The dual problem of SVM can be written in matrix format as shown in following,
\begin{equation}\label{eq:svm_dual_matrix}
\begin{split}
\max \ \ \ & \mathbf{1}^{T}\boldsymbol{\alpha} - \frac{1}{2}\boldsymbol{\alpha}^{T}\mathbf{Y}\mathbf{X}\mathbf{X}^{T}\mathbf{Y}\boldsymbol{\alpha} \\
s.t \ \ \ & \mathbf{1}^{T}\mathbf{Y}\boldsymbol{\alpha} = \mathbf{0} \\
\ \ \ & \mathbf{0} \leq \boldsymbol{\alpha} \leq \mathbf{C},
\end{split}
\end{equation}
where $\mathbf{X} \in \mathbb{R}^{n \times d}$ is the input feature matrix, and $\mathbf{Y}$ is a diagonal matrix where the elements in the diagonal are class labels, $\mathbf{Y}_{ii} = y_{i}$, $C$ is the regularization parameter.

By applying SRHT to project original $\mathbf{X}$ from $d$-dimensional space into $r$-dimensional space, i.e., $\widetilde{\mathbf{X}} = \sqrt{\frac{d}{r}}\mathbf{XDHS}$, the SVM optimization problem on projected data $\widetilde{\mathbf{X}}$ will be
\begin{equation}\label{eq:svm_dual_matrix_with_SHRT}
\begin{split}
\max \ \ \ & \mathbf{1}^{T}\boldsymbol{\widetilde{\alpha}} - \frac{1}{2}\boldsymbol{\widetilde{\alpha}}^{T}\mathbf{Y}\widetilde{\mathbf{X}}\widetilde{\mathbf{X}}^{T}\mathbf{Y}\boldsymbol{\widetilde{\alpha}} \\
s.t \ \ \ & \mathbf{1}^{T}\mathbf{Y}\boldsymbol{\widetilde{\alpha}} = \mathbf{0} \\
\ \ \ & \mathbf{0} \leq \boldsymbol{\widetilde{\alpha}} \leq \mathbf{C}.
\end{split}
\end{equation}

Comparing the dual problem (\ref{eq:svm_dual_matrix}) in the original space and the dual problem (\ref{eq:svm_dual_matrix_with_SHRT}) in the projected low-dimensional space, the only difference is that $\mathbf{X}\mathbf{X}^{T}$ in (\ref{eq:svm_dual_matrix}) is replaced by $\widetilde{\mathbf{X}}\widetilde{\mathbf{X}}^{T}$ in (\ref{eq:svm_dual_matrix_with_SHRT}). The constraints on optimal solutions (i.e., $\boldsymbol{\alpha}^*$ of (\ref{eq:svm_dual_matrix}) and $\boldsymbol{\widetilde{\alpha}}^*$ of (\ref{eq:svm_dual_matrix_with_SHRT}) ) do not depend on input data matrix. Therefore, $\boldsymbol{\widetilde{\alpha}}^*$ is expected to be close to $\boldsymbol{\alpha}^*$ when $\widetilde{\mathbf{X}}\widetilde{\mathbf{X}}^{T}$ is close to $\mathbf{X}\mathbf{X}^{T}$.

Let us rewrite $\widetilde{\mathbf{X}} = \mathbf{X}_r\mathbf{S}\sqrt{\frac{d}{r}}$ where $\mathbf{X}_r = \mathbf{XDH}$, we can view the effect of $\mathbf{X}_r\mathbf{S}\sqrt{\frac{d}{r}}$ is to uniformly sample $r$ columns from $\mathbf{X}_r$ and then re-scale each selected column by a scaling factor $\sqrt{\frac{d}{r}}$. It corresponds to use uniform column sampling for approximating matrix multiplication \cite{mahoney2011randomized}. It has been proved in \cite{drineas2006fast} that ($i,j$)-th element in $\widetilde{\mathbf{X}}\widetilde{\mathbf{X}}^{T}$ equals to the ($i,j$)-th element of exact product $\mathbf{X}_r\mathbf{X}_r^{T}$ in expectation regardless of the sampling probabilities. However, the variance of the ($i,j$)-th element in $\widetilde{\mathbf{X}}\widetilde{\mathbf{X}}^{T}$ depends on the choice of sampling probabilities. Therefore, simply using uniform sampling would result in large variance of each element in $\widetilde{\mathbf{X}}\widetilde{\mathbf{X}}^{T}$ and then cause a large approximation error between $\widetilde{\mathbf{X}}\widetilde{\mathbf{X}}^{T}$ and $\mathbf{X}_r\mathbf{X}_r^{T}$.

Motivated by this observation, we first propose to improve SRHT by employing importance sampling for randomized matrix multiplication. Specifically, we would like to design a method to construct $\widetilde{\mathbf{S}}$ based on specific data properties of $\mathbf{X}_r$ instead of using random sampling matrix $\mathbf{S}$ in original SRHT. The idea is based on importance sampling for approximating matrix multiplication: (1) we assign each column $i$ in $\mathbf{X}_r$ a probability $p_{i}$ such that $p_{i} \ge 0$ and $\sum_{i=1}^{d}p_i = 1$; (2) we then select $r$ columns from $\mathbf{X}_r$ based on the probability distribution $\{p_{i}\}_{i=1}^{d}$ to form $\widetilde{\mathbf{X}}$; (3) a column in $\widetilde{\mathbf{X}}$ is formed as $\sqrt{\frac{1}{rp_{i}}}\mathbf{X}_{r({:,i}})$ if this column is chosen from the $i$-th column in $\mathbf{X}_r$ where $\sqrt{\frac{1}{rp_{i}}}$ is the re-scaling factor. In other word, the matrix $\widetilde{\mathbf{S}} \in \mathbb{R}^{d \times r}$ is constructed as: for every column $j$ in $\widetilde{\mathbf{S}}$ only one entry $i$ ($i$ is a random number from $\{1, \dots, d\}$) is selected by following the probability distribution $\{p_{i}\}_{i=1}^{d}$ and a non-zero value is assigned to this entry. The non-zero value is set as
\begin{eqnarray}\label{def:S}
\widetilde{\mathbf{S}}_{ij} = \sqrt{\frac{1}{rp_{i}}}.
\end{eqnarray}

In the uniform sampling setting, $p_{i} = \frac{1}{d}$, therefore $\widetilde{\mathbf{S}} =  \sqrt{\frac{d}{r}}\mathbf{S}$ which is consistent with the setting in original SRHT as shown in (\ref{eq:SRHT}).

Now, the question is what choice of probability $\{p_{i}\}_{i=1}^{d}$ can optimally reduce the expected approximation error between $\widetilde{\mathbf{X}}\widetilde{\mathbf{X}}^{T}$ and $\mathbf{X}\mathbf{X}^{T}$. We use Frobenius norm to measure the approximation error between $\widetilde{\mathbf{X}}\widetilde{\mathbf{X}}^{T}$ and $\mathbf{X}\mathbf{X}^{T}$.

Let us use $\mathbf{X}_{r(:,j)}$ to denote the $j$-th column of $\mathbf{X}_r$, as shown in the Proposition \ref{unsupervised_optimal_probaility}, the optimal sampling probability to minimize $\mathbf{E}[\|\widetilde{\mathbf{X}}\widetilde{\mathbf{X}}^{T} - \mathbf{X}\mathbf{X}^{T}\|_F]$ is

\begin{eqnarray}\label{optimal_p}
p_j = \frac{\|\mathbf{X}_{r(:,j)}\|_2^{2}}{\sum_{j=1}^{d}\|\mathbf{X}_{r(:,j)}\|_2^{2}}.
\end{eqnarray}

\begin{proposition}\label{unsupervised_optimal_probaility}
Suppose $\mathbf{X} \in \mathbb{R}^{n \times d}$, $\mathbf{X}_r = \mathbf{XDH} \in \mathbb{R}^{n \times d}$ and $\widetilde{\mathbf{X}} = \mathbf{X}_r\widetilde{\mathbf{S}}_{ij}$ where $\widetilde{\mathbf{S}}_{ij} \in \mathbb{R}^{d \times r}$ is defined as in (\ref{def:S}). Then, the optimal sampling probability to minimize the expectation of the Frobenius norm of the approximation error between  $\widetilde{\mathbf{X}}\widetilde{\mathbf{X}}^{T}$ and $ \mathbf{X}\mathbf{X}^{T}$, i.e.,  $\mathbf{E}[\|\widetilde{\mathbf{X}}\widetilde{\mathbf{X}}^{T} - \mathbf{X}\mathbf{X}^{T}\|]$, is $p_j = \frac{\|\mathbf{X}_{r(:,j)}\|_2^{2}}{\sum_{j=1}^{d}\|\mathbf{X}_{r(:,j)}\|_2^{2}}$.
\end{proposition}

To prove Proposition \ref{unsupervised_optimal_probaility}, we first use the fact $\mathbf{X}\mathbf{X}^T = \mathbf{X}_r\mathbf{X}_r^{T}$ since $\mathbf{D}\mathbf{H}\mathbf{H}^{T}\mathbf{D}^{T} = \mathbf{I}$. Then, we apply the Lemma 4 from \cite{drineas2006fast} to obtain the optimal sampling probability. 

In addition, we also propose a deterministic top-$r$ sampling, which select the $r$ columns with largest Euclidean norms from $\mathbf{X}_r$. As shown in the Proposition \ref{proposition:upperbound}, this deterministic top-$r$ sampling method directly optimizes an upper bound of  $\|\widetilde{\mathbf{X}}\widetilde{\mathbf{X}}^{T} - \mathbf{X}\mathbf{X}^{T}\|_F$ without using re-scaling factor.

\begin{proposition}\label{proposition:upperbound}
Suppose $\mathbf{X} \in \mathbb{R}^{n \times d}$, $\mathbf{X}_r = \mathbf{XDH} \in \mathbb{R}^{n \times d}$ and $\widetilde{\mathbf{X}} = \mathbf{X}_r\mathbf{P}$ where $\mathbf{P} \in \mathbb{R}^{d\times d}$ is a diagonal matrix whose elements are 1 or 0. $\mathbf{P}_{ii} = 1$ denotes the $i$-th column in $\mathbf{X}_r$ is selected. Then, $\sum_{\mathbf{P}_{jj} = 0} \|\mathbf{X}_{r(:,j)}\|_{2}^2$ is an upper bound of the approximation error $\|\widetilde{\mathbf{X}}\widetilde{\mathbf{X}}^{T} - \mathbf{X}\mathbf{X}^{T}\|_F$.
\end{proposition}
\proof
Since $\widetilde{\mathbf{X}} = \mathbf{X}_r\mathbf{P}$ and $\mathbf{X}\mathbf{X}^T = \mathbf{X}_r\mathbf{X}_r^{T}$,
$\|\widetilde{\mathbf{X}}\widetilde{\mathbf{X}}^{T} - \mathbf{X}\mathbf{X}^{T}\|_F = \|\mathbf{X}_r\mathbf{P}\mathbf{P}^T\mathbf{X}_r^{T} - \mathbf{X}_r\mathbf{X}_r^{T} \|_F$. Let us rewrite the matrix product $\mathbf{X}_r\mathbf{X}_r^{T}$ as the sum of $d$ rank one matrices $\sum_{j=1}^{d}\mathbf{X}_{r(:,j)}{\mathbf{X}_{r(:,j)}}^{T}$. Similarly, $\mathbf{X}_r\mathbf{P}\mathbf{P}^T\mathbf{X}_r^{T}$ can be rewritten as $\sum_{j=1}^{d}{\mathbf{P}_{jj}\mathbf{X}_{r(:,j)}{\mathbf{X}_{r(:,j)}}^{T}}$. Therefore, the approximation error $\|\widetilde{\mathbf{X}}\widetilde{\mathbf{X}}^{T} - \mathbf{X}\mathbf{X}^{T}\|_F$ can be upper bounded as follows,
\begin{equation}
\label{eq:gap_upperbound}
\begin{split}
& \|\widetilde{\mathbf{X}}\widetilde{\mathbf{X}}^{T} - \mathbf{X}\mathbf{X}^{T}\|_F
=\|\sum_{\mathbf{P}_{jj} = 0} \mathbf{X}_{r(:,j)}{\mathbf{X}_{r(:,j)}}^{T}\|_{F} \\
\le & \sum_{\mathbf{P}_{jj} = 0} \|\mathbf{X}_{r(:,j)}{\mathbf{X}_{r(:,j)}}^{T}\|_{F}
= \sum_{\mathbf{P}_{jj} = 0} \|\mathbf{X}_{r(:,j)}\|_{2}^2
\end{split}
\end{equation}

In our experimental section, we have shown that this deterministic top-$r$ sampling obtains better classification accuracy than importance sampling method. \citeauthor{adelman2018faster}\cite{adelman2018faster} also observed that deterministic top-$r$ sampling gets better results when they use randomized matrix multiplication approximation to speedup the training of deep neural networks.

\subsection{Improved SRHT by Supervised Non-uniform Sampling}
In this section, we describe our proposed non-uniform sampling method from supervised perspective. We propose to construct the sampling matrix $\widetilde{\mathbf{S}}$ by incorporating label information. Our proposed method is based on the idea of metric learning \cite{xing2003distance}. The idea is to select $r$ columns from $\mathbf{X}_r$ for improving the inter-class separability along with the intra-class compactness as used in \cite{lan19scaling}.

Let $\mathbf{z}_i$ be new feature representation of the $i$-th data sample in $\mathbf{X}_r$. Then, we measure the interclass separability by the sum of squared pair-wise distances between $\mathbf{z}_i$'s from different classes as $\sum _{y_i \neq y_j} \|\mathbf{z}_i - \mathbf{z}_j\|^2$. Similarly, we measure the intra-class tightness by the sum of squared pair-wise distances between $\mathbf{z}_i$'s from the same class as $\sum _{y_i = y_j} \|\mathbf{z}_i - \mathbf{z}_j\|^2$. We formulate our objective as a combination of these two terms,
\begin{equation}\label{inter-intra}
\begin{split}
  &\sum\limits_{i,j \in {D}} \|\mathbf{z}_i - \mathbf{z}_j\|^2\mathbf{A}_{ij}
= trace((\X_{r}\widetilde{\mathbf{S}})^\top\mathbf{L}(\X_{r}\widetilde{\mathbf{S}})) \\
&= trace({\X_{r}}^\top\mathbf{L}\X_{r}\widetilde{\mathbf{S}}\widetilde{\mathbf{S}}^\top),
\end{split}
\end{equation}
	where $\mathbf{A}_{ij} = 1$ if $y_i = y_j$, and $\mathbf{A}_{ij} = -a$ if $y_i \neq y_j$, parameter $a$ is used to balance the tradeoff between inter-class separability and intra-class tightness, $\mathbf{L}$ is the Laplacian matrix of $\mathbf{A}$, defined as $\mathbf{L} = \mathbf{D} - \mathbf{A}$, and $\mathbf{D}$ is a diagonal degree matrix of $\mathbf{A}$ such that $\mathbf{D}_{ii} = \sum_{j}\mathbf{A}_{ij}$. $\widetilde{\mathbf{S}}$ is the sampling matrix whose elements are 0 or 1. We want to learn $\widetilde{\mathbf{S}}$ by minimizing (\ref{inter-intra}).

Without losing any information, let us rewrite our variable $\widetilde{\mathbf{S}}$ as a $d \times d$ diagonal matrix $\mathbf{P}$ where the diagonal element $\mathbf{P}_{ii} \in \{0, 1\}$. $\mathbf{P}_{ii} = 1$ means the $i$-th column in $\mathbf{X}_r$ is selected and $\mathbf{P}_{ii} = 0$ means the $i$-th column is not selected. Then $\widetilde{\mathbf{S}}$ can be viewed as a submatrix of $\mathbf{P}$ that selects the columns with $\mathbf{P}_{ii} = 1$ from $\mathbf{P}$.

It is straight-forward to verify that $\widetilde{\mathbf{S}}\widetilde{\mathbf{S}}^\top$ in (\ref{inter-intra}) equals to $\mathbf{P}$. Therefore, our optimization is formulated as
\begin{equation}\label{eq:metricLearningObj}
\begin{split}
\min \limits_{\boldsymbol{\mathbf{P}}}&\ \ \ trace({\X_{r}}^\top\mathbf{L}\X_{r}\mathbf{P}) \\
	s.t& \ \boldsymbol{P}_{ii} \in \{0,1\}, \\
       & \ \sum_{i=1}^{d}\boldsymbol{P}_{ii} = r.
\end{split}
\end{equation}
	
	Let vector $\mathbf{b}$ denote the diagonal of the matrix ${\X_{r}}^\top\mathbf{L}\X_{r}$, i.e $b_i = ({\X_{r}}^\top\mathbf{L}\X_{r})_{ii}$, vector $\boldsymbol{p}$ denote the diagonal of $\mathbf{P}$. Then, the objective (\ref{eq:metricLearningObj}) is simplified to
\begin{equation}\label{eq:metricLearningObj3}
	\begin{split}
	\min\limits_{\boldsymbol{p}}& \ \ \ \ \mathbf{b}^\top\boldsymbol{p} \\
	\text{s.t.} & \ \ \ \ p_{i} \in \{0, 1\}, \\
&\sum_{i=1}^{d}p_{i} = r.
	\end{split}
	\end{equation}

The optimal solution of (\ref{eq:metricLearningObj3}) is just to select the largest $r$ elements in $\mathbf{b}_i$ and set the corresponding entries in $\mathbf{p}_i$ to 1. Then $\widetilde{\mathbf{S}}$ can easily obtained based on non-zero values in $\mathbf{p}$.

\begin{algorithm}[tb]
\caption{Improved Subsampled Randomized Hadamard Transform (ISRHT) for linear SVM classification}
\begin{algorithmic}

		\STATE \underline{\textbf{Training}}
		\STATE \textbf{Input}: training set $\textbf{X}_{train}$, reduced dimension $r$, regularization parameter $c$;
		\STATE \textbf{Output}: model $\mathbf{w}\in\mathbb{R}^{r\times 1}$, matrix $\mathbf{D} \in \mathbb{R}^{d \times d}$ and $\widetilde{\textbf{S}} \in \mathbb{R}^{d \times r}$;
	\end{algorithmic}
	\begin{algorithmic}[1]
		\STATE Generate a diagonal random sign matrix $\mathbf{D}$ as in (\ref{eq:SRHT})
		\STATE Compute $\mathbf{X}_{r} = \mathbf{X}_{train}\mathbf{DH} \in \mathbb{R}^{n \times d}$ by FFT
		\STATE Obtain the sampling matrix $\widetilde{\textbf{S}}$ 
		\STATE Compute $\hat{\mathbf{X}}_{train} = \mathbf{X}_{r}\widetilde{\textbf{S}} \in \mathbb{R}^{n \times r}$
		\STATE Train a linear SVM on $\hat{\mathbf{X}}_{train}$ and obtain the model $\mathbf{w}$
	\end{algorithmic}
	\begin{algorithmic}
		\STATE \underline{\bf{Prediction}} \STATE
		\textbf{Input}: test data $\mathbf{X}_{test}$, model $\mathbf{w}$, matrix $\mathbf{D}, \widetilde{\textbf{S}}$;
		\STATE \textbf{Output}: predicted labels $\hat{\mathbf{y}}_{test}$;
		
	\end{algorithmic}
	\begin{algorithmic}[1]
		\STATE compute $\hat{\mathbf{X}}_{test} = \mathbf{X}_{test}\mathbf{DH}\widetilde{\textbf{S}} \in \mathbb{R}^{n \times r}$
		\STATE predict by $\hat{\textbf{y}}_{test} = \mathbf{\hat{X}}_{test}\mathbf{w}$;
	\end{algorithmic}
    \label{alg:ISRHT}
\end{algorithm}

\subsection{Algorithm Implementation and Analysis}
We summarize our proposed algorithm in algorithm \ref{alg:ISRHT} and name it as Improved Subsampled Randomized Hadamard Transform (ISRHT). With respect to training time complexity, step 2 needs $O(ndlog(d))$ time by using FFT. In step 3, for unsupervised column selection, we need $O(nd)$ time to compute the Euclidean norm for each column. For supervised column selection, $b_i$ in (\ref{eq:metricLearningObj3}) are computed as $b_i = ({\X_{r}}^\top\mathbf{L}\X_{r})_{ii}$ and the Laplacian matrix \textbf{L} is equal to $\mathbf{D - yy^T}$, therefore, the required time is still $O(nd)$. Step 4 needs $O(r)$ time and step 5 needs $O(tnr)$ where $t$ is the number of iterations needed for training linear SVM. Therefore, the overall time complexity of our proposed algorithm is $O(ndlog(d) + nd + tnr)$. With respect to prediction time complexity, our proposed method needs $O(d\log(d)+r)$ time to make a prediction for a new test sample. With respect to space complexity, $\mathbf{D}$ is a diagonal matrix, matrix $\widetilde{\mathbf{S}}$ is a sampling matrix with $r$ non-zero entries. Therefore, the required space for storing the projection matrix $\mathbf{R}$ of our proposed methods is $O(d+r)$. Our proposed methods achieve very efficient time and space complexities for model deployment. We compare the time and space complexity of different random projection methods in Table \ref{time_space_different_algorithms}.

\begin{table}[tb]
		\centering
		\caption{Comparison of different random projection methods}
		\begin{tabular}{|c|c|c|}
		\hline
        Algorithms & Space for $\mathbf{R}$ & Time for projection \\
        \hline
	    Gaussian  & $O(dr)$  & $O(ndr)$ \\
        \hline
        Achlioptas & $O(\frac{1}{3}dr)$ & $O(\frac{1}{3}ndr)$ \\
        \hline
        Sparse Embedding  & $O(d)$  & $O(nnz(\mathbf{X}))$  \\
        \hline
        SRHT  & $O(d + r)$ & $O(ndlog(r))$ \\
        \hline
        ISRHT-unsupervised  & $O(d+r)$ & $O(ndlog(d) + nd)$  \\
        \hline
        ISRHT-supervised & $O(d+r)$ & $O(ndlog(d) + nd)$  \\
		\hline
		
	\end{tabular}
    \label{time_space_different_algorithms}
\end{table}

\subsection{Extension to High-dimensional Sparse Data}

In this section, we describe an extension of our algorithm on high-dimensional sparse data. One disadvantage of  applying Hadamard transform on sparse data is that it produces a dense matrix. As shown in the step 2 in Algorithm \ref{alg:ISRHT}, the memory cost is increased from $O(nnz(\mathbf{X}))$ to $O(nd)$. To tackle this challenge, we use a combination of sparse embedding and SRHT as proposed in \cite{chen2015fast} for memory-efficient projection. We improve their work by replacing the original SRHT by our proposed ISRHT methods. We expect that our proposed data-dependent projection methods can obtain higher accuracies than data-independent projection methods on high-dimensional sparse data.

Specifically, we first use sparse embedding matrix $\mathbf{R}_{sparse} \in \mathbb{R}^{d \times r^{\prime}}$ to project $\mathbf{X}$ into $r^{\prime}$-dimensional space and then use our improved SRHT methods to project into $r$ dimensional space. The sparse embedding matrix $\mathbf{R}_{sparse}$ is generated as follows, for each row in random matrix $\mathbf{R}_{sparse}$, randomly selected one column and assign either 1 or $-1$ with probability 0.5 to this entry. All other entries are 0s. Due to the sparse structure of $\mathbf{R}_{sparse}$, matrix multiplication $\mathbf{X}\mathbf{R}_{sparse}$ takes $O(nnz(\mathbf{X}))$ time. By doing this, the time complexity of the proposed method on high-dimensional sparse data was reduced to $O(nnz(\mathbf{X}) + nr^{\prime}log(r^{\prime}) + nr^{\prime})$ and the memory cost was reduced to $O(nnz(\mathbf{X}) + nr^{\prime})$. In our experimental setting, we set $r^{\prime} = 2r$ as suggested in \cite{chen2015fast}.

\begin{table*}[tbh]

	\centering
	\caption{The classification accuracy and running time (in seconds) (The best results for each dataset are in bold and italic and the best results among the seven random projection methods are in bold).}
	\begin{tabular}{c|c|c|c|c|c|c}
		\hline
		\diagbox{Algorithms}{Datasets}	  & {\sf mushrooms}  &  {\sf usps-b}       & {\sf gisette}     & {\sf real-sim}    & {\sf rcv1-binary}        & {\sf news20-binary} \\
		          & ($r$ = 16) &  ($r$ = 16) & ($r$ = 256) & ($r$ = 256) & ($r$ = 256) & ($r$ = 256) \\
		\hline
		\multirow{2}{*}{All features} &  \textbf{\textit{99.85}}      & \textbf{\textit{91.03}}   & \textbf{\textit{97.5}}        & \textbf{\textit{97.43}}      & \textbf{\textit{96.43}}    & \textbf{\textit{96.01}}          \\
		             &  0.1s       & 2.8s    & 3.0s        & 0.1s       &0.1s     & 0.3s          \\
		\hline
		\multirow{2}{*}{PCA}         & 96.23        & 85.85   & 96.50        & 93.10      &95.05    & -          \\
		            & 0.3s         & 0.6s    & 3.2s        & 51.0s      &34.5s    & -         \\
		 \hline
		 \multirow{2}{*}{DLSS}    & 94.66& 80.21 & 93.90 & 89.89 &  91,43& 79.33\\
		            & 0.2s         & 1.0s           &  5.8s
		            &        2.0s    &  1.6s & 9.3s      \\
		\hline
		\hline
		           
		\multirow{2}{*}{Gaussian}    & 89.15$\pm$3.85 & 80.05$\pm$3.33 & 89.76$\pm$0.56 & 76.48$\pm$0.55 & 78.90$\pm$0.61 &70.10$\pm$0.78 \\
		            & 0.2s          & 1.0s            &5.9s             &19s             &1.2s    &20.3s          \\
		\hline
		\multirow{2}{*}{Achlioptas}  & 91.74$\pm$2.75 & 81.50$\pm$2.28 & 89.56$\pm$0.65 & 76.20$\pm$0.63 & 79.03$\pm$1.28 & 69.29$\pm$0.99          \\
		            & 0.3s          & 0.9s             &7.8s             &6.3s             &1.5s  & 28.6s          \\
		\hline
		\multirow{2}{*}{Sparse Embedding}      & 89.06$\pm$4.18  & 82.09$\pm$2.05 & 89.65$\pm$0.86 & 76.82$\pm$0.45 & 79.11$\pm$1.20 & 70.08$\pm$0.69          \\
		   & 0.3s           & 0.8s             & 1.5s             & 2.4s             & 0.2s & 0.6s          \\
		\hline
		\multirow{2}{*}{SRHT}        & 92.45$\pm$2.59  & 81.87$\pm$1.55 &  91.77$\pm$0.76 & 72.80$\pm$0.23 & 79.08$\pm$0.97 & 70.58$\pm$0.26          \\
		            & 0.3s        & 0.3s            & 1.5s             & 23.3s            & 0.5s  &0.7s          \\
		\hline
		\multirow{2}{*}{ISRHT-nps}      & 94.30$\pm$1.15 & 83.34$\pm$1.31 & 91.17$\pm$0.51 & 73.04$\pm$0.51 & 79.28$\pm$0.63 & 70.36$\pm$0.84   \\
		         & 0.4s         & 0.6s            & 1.4s             & 24.2s            & 0.6s            &0.9s          \\
		\hline
		\multirow{2}{*}{ISRHT-top-$r$}      &94.23$\pm$1.58 &82.91$\pm$1.07  &93.20$\pm$0.61 &76.81$\pm$0.46 & 82.81$\pm$0.89 & 73.46$\pm$0.90 \\
		       &0.4s         & 0.4s          &1.4s             &18.7s             & 0.6s            &0.9s          \\
		\hline
		\multirow{2}{*}{ISRHT-supervised}      &\textbf{96.25$\pm$1.22} & \textbf{86.25$\pm$0.81} & \textbf{93.90$\pm$0.72} & \textbf{80.04$\pm$0.52} & \textbf{87.81$\pm$0.42} & \textbf{78.33$\pm$0.74}          \\
		  &0.2s     & 0.4s            &1.4s            &44.3s             &1.0s             & 1.1s         \\
		\hline
	\end{tabular}\label{experimentalResult}
\end{table*}

\section{Experiments}
In this section, we compare our proposed methods with other four popular random projection methods on six real-world benchmark datasets. These benchmark datasets are downloaded from \textbf{LIBSVM} website \footnote{\url{https://www.csie.ntu.edu.tw/~cjlin/libsvmtools/datasets/}}. For datasets ({\sf mushrooms},{\sf real-sim}, {\sf rcv1-binary}, {\sf news20-binary}), we refer to the setting in \cite{pourkamali2018randomized} and randomly select 70\% of the original training data as training data and the rest 30\% as test data. The detailed information about these six datasets is summarized in in Table \ref{table.data}.
\begin{table}[htb]
	\centering
	\caption{Experiment dataset}
	\begin{tabular}{cccc}
		\hline
		Dataset & train size & test size & dimensionality     \\
		\hline
		{\sf mushrooms} & 6,000	& 2124  &  112 \\			
		{\sf usps-b}	  & 7,291	& 2007 &  256    \\
		{\sf gisette}   & 6,000   & 1000 &  5,000  \\
		{\sf real-sim}  & 48,447 	& 23862 &  20,958\\
		{\sf rcv1-binary}      & 13,562  & 6680 &  47,236 \\
		{\sf news20-binary}    & 13,397  & 6599 &  1,355,191\\
		\hline
		\label{table.data}
	\end{tabular}	
\end{table}

We evaluate the performance of the following seven random projection algorithms in our experiments:
\begin{itemize}[leftmargin=*]
\item Gaussian: random projection by Gaussian Matrix \cite{dasgupta1999elementary};
\item Achlioptas: random projection by Achlioptas matrix \cite{achlioptas2003database};
\item Sparse Embedding: random projection by count sketch matrix \cite{clarkson2017low};
\item SRHT: original subsampled randomized Hadamard transform with uniform sampling \cite{tropp2011improved};
\item ISRHT-nps: our proposed method for ISRHT with norm-proportional sampling as defined in (\ref{optimal_p});
\item ISRHT-top-$r$: our proposed deterministic sampling method for ISRHT which select $r$ columns with largest Euclidean norms;
\item ISRHT-supervised: our proposed method by incorporating label information as defined in (\ref{eq:metricLearningObj3});
\end{itemize}
We also include the results of using all features and two other popular data-dependent dimensionality reduction methods: PCA and Deterministic Leverage Score Sampling (DLLS) \cite{papailiopoulos2014provable} in Table \ref{experimentalResult}. Due to high computational complexity of Singular Value Decomposition (SVD) in PCA and leverage score sampling, we employed randomized SVD \cite{halko2011finding} in our experiments.

\textbf{Experimental setup}. The feature values for all data sets are linearly scaled to [$-1$, 1]. For dense datasets ({\sf mushrooms}, {\sf usps-b} and {\sf gisette}), we directly apply our proposed ISRHT methods on them. For the high-dimefnsional sparse datasets ({\sf real-sim}, {\sf rcv1-binary} and {\sf news20-binary}), we use an extension of our proposed methods that combines of sparse embedding and ISRHT for memory efficient projection as discussed in previous section. The regularization parameter $C$ in linear SVM is chosen from \{$2^{-5}, 2^{-4}, \dots, 2^{4}, 2^{5}$\} by 5-fold cross validation. The tradeoff parameter $a$ for ISRHT-supervised is fixed to 1.0. Our experiments are performed on a server with Dual 6-core Intel Xeon 2.4GHz CPU and 128 GB RAM.
\begin{figure*}[tbh]
		\centering	
         \subfigure[mushrooms]{
			\label{Fig.mushrooms}
			\includegraphics[width=0.62\columnwidth]{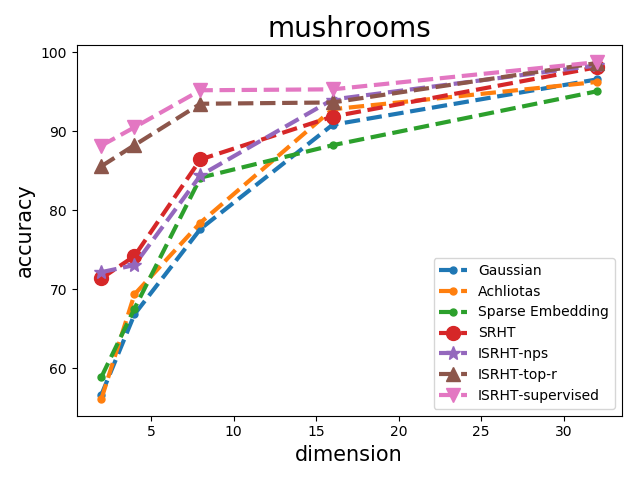}}
         \subfigure[usps]{
			\label{Fig.usps}
			\includegraphics[width=0.62\columnwidth]{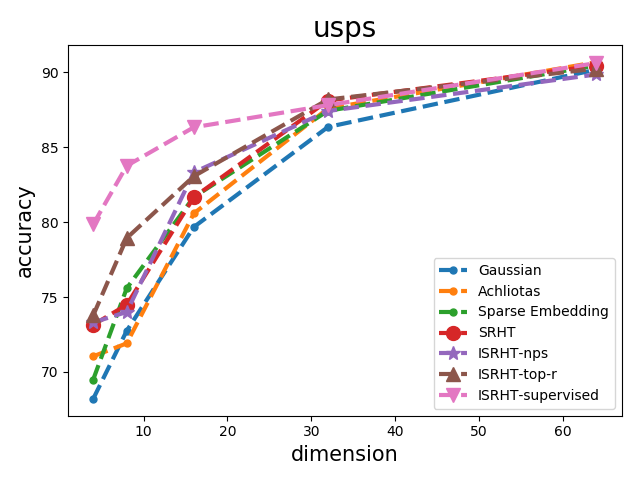}}	
		\subfigure[gisette]{
			\label{Fig.gisette}
			\includegraphics[width=0.62\columnwidth]{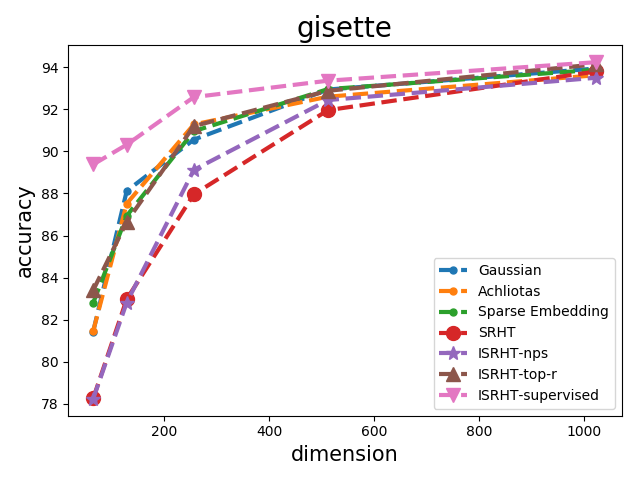}} \\
        \subfigure[real-sim]{
			\label{Fig.real-sim}
			\includegraphics[width=0.62\columnwidth]{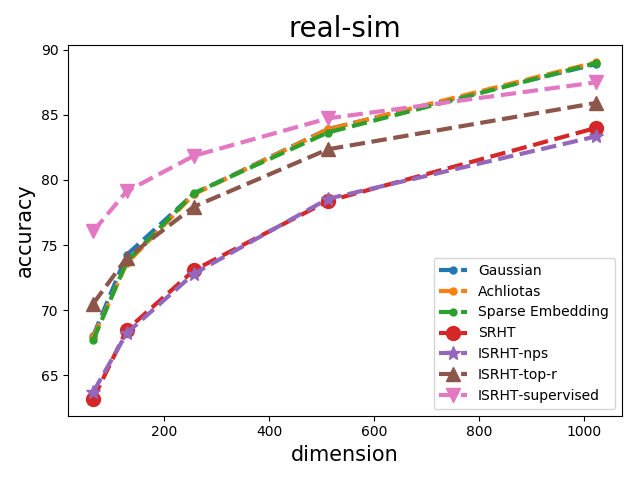}}	\subfigure[rcv]{
			\label{Fig.rcv}
			\includegraphics[width=0.62\columnwidth]{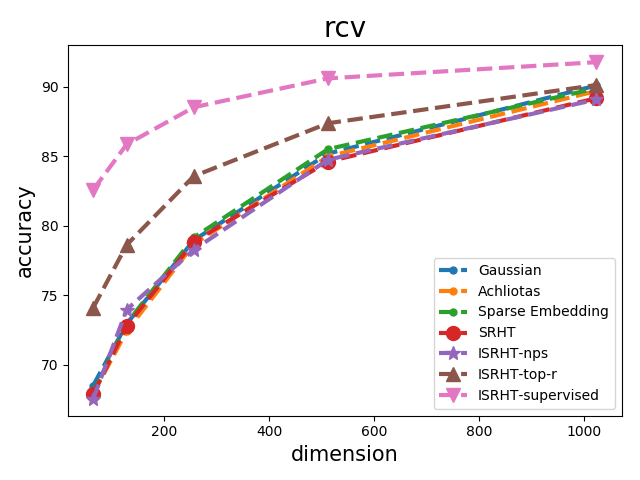}}
        \subfigure[news]{
			\label{Fig.news}			
			\includegraphics[width=0.62\columnwidth]{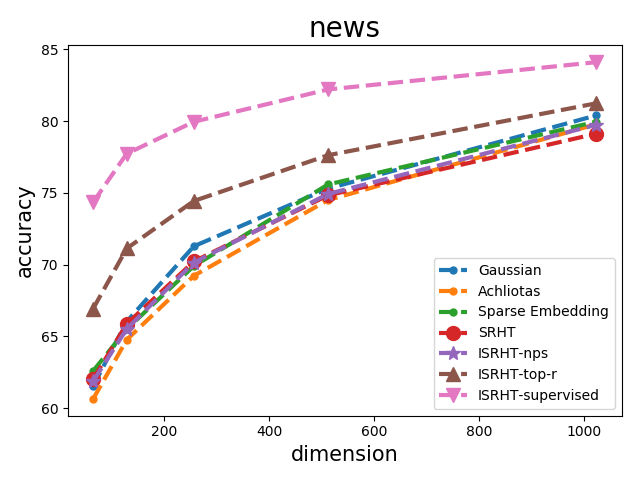}}
		
       \caption{classification accuracies of different algorithms with different $r$}
       \label{fig:results_wrt_r}
\end{figure*}

\textbf{Experimental Results}. The accuracy and training time of all algorithms are reported in Table \ref{experimentalResult}. '-' is used to denote that PCA cannot be completed because of space complexity. The number of reduced dimension $r$ is shown in the first column of the table. We also explore the impact of parameter $r$ in Figure \ref{fig:results_wrt_r} and will be discussed later. The reported accuracies are the averaged accuracies on test data based on 15 repetitions. As shown in Table \ref{experimentalResult}, our proposed ISRHT-(top-$k$) method achieves significant higher accuracies than other four data-independent random projection methods (i.e., Gaussian, Achlioptas, Sparse Embedding and SRHT) on all six datasets. The ISRHT-nps gets slightly better results than SRHT. Furthermore, by incorporating the label information, our proposed ISRHT-supervised method gets the best accuracy on all six datasets among the seven random projection methods. Using all features gets the best accuracy on all six datasets. PCA gets better accuracy than DLSS and other random projection methods while needs longer running time. We also observe that our proposed methods produce dense feature representation after random projection and therefore it needs longer running time than directly applying liblinear on high-dimensional sparse data. 

With respect to the running time, all the random projection methods are more efficient than data-dependent dimension reduction method such as PCA and DLSS especially on large datasets. Among these random projection methods Gaussian and Achiloptas are much slower than Sparse Embedding and SRHT since they need $O(ndr)$ time for projection. The running times of our proposed methods ISRHT-nps and ISRHT-top-r are very close to SRHT. The results demonstrate that our proposed methods only slightly increase the running time since the computing norm can be done very efficiently. 

\textbf{Impact of parameter $r$}. We evaluate the impact of parameter $r$ in our proposed algorithms. We show the accuracies of all the algorithms with respect to different $r$ in Figure \ref{fig:results_wrt_r}. We can observe that our proposed methods gets higher accuracy than other four methods. The accuracy improvement is large when the parameter $r$ is set to relative small number. As expected, the difference will become smaller as the parameter $r$ increases.     
\begin{figure}[tbh]
		\centering
		\subfigure[ISRHT-nps vs. SRHT]{\label{Fig.parameter.1}
			\includegraphics[width=0.225\textwidth]{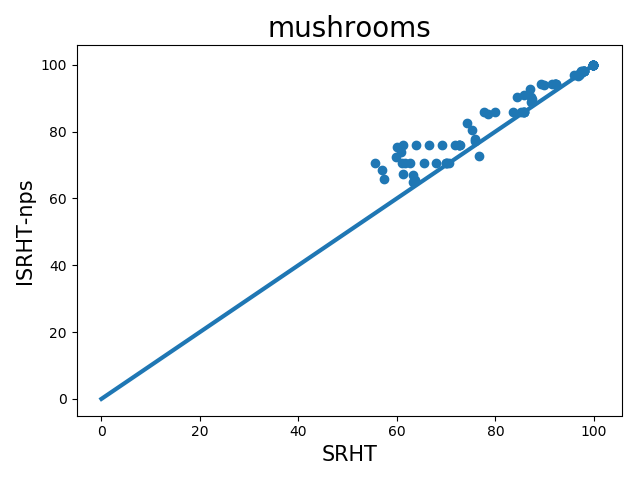}}
		\subfigure[ISRHT-top-$r$ vs. SRHT]{\label{Fig.parameter.2}
			\includegraphics[width=0.225\textwidth]{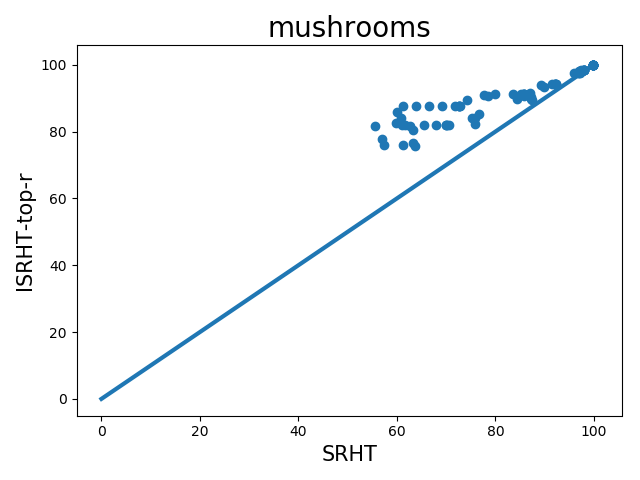}}\\
		\caption[]{Classification Accuracies of SRHT (x-axis) and ISRHT (y-axis) for different parameters.}
		\label{fig:hyperparameter}
\end{figure}

\textbf{Stability Comparison Between ISRHT and SRHT}. In this section, we would like to further investigate the stability of our proposed methods and SRHT with respect to different choices of the regularization parameter $C$ in SVM and reduced dimension $r$. We evaluate all methods for different parameter combinations of $C$ and $r$. Regularization parameter $C$ is chosen from $\{2^{-5}, 2^{-4}, \dots, 2^{5}\}$ and reduced dimension $r$ is chosen from $\{\frac{d}{2^6},\frac{d}{2^5},\dots,\frac{d}{2}\}$.

Due to the space limitation, we only show the results of comparing the prediction accuracies of ISRHT-nps and ISRHT-top-$r$ with SRHT on {\sf mushrooms} dataset in Figure $\ref{fig:hyperparameter}$. The accuracies plotted in Figure $\ref{fig:hyperparameter}$ are based on the average of 15 repetitions. For each parameter combination, we plot its corresponding accuracy of SRHT on the $x$-axis and the accuracy of our proposed methods on $y$-axis. So the points above $y = x$ line indicate an accuracy improvement of our proposed methods (y-axis) (ISRHT-nps or ISRHT-top-$r$) over SRHT(x-axis). Overall speaking, our proposed methods is more stable with different choices of parameter combinations.

\section{Conclusion and Future work}
In this paper, we propose to produce more effective low-dimensional embedding than original SRHT by using non-uniform sampling instead of uniform sampling in SRHT. To achieve this goal, we first analyze the effect of using SRHT for random projection in the context of linear SVM classification. Based on our analysis, we have proposed importance sampling and deterministic top-$r$ sampling to improve the embedding. Secondly, we also propose a new sampling method to incorporate label information based on the idea of metric learning. We performed extensive experiments to evaluate our proposed non-uniform samplings methods. Our experimental results demonstrate that our proposed new methods can achieve better accuracy than original SRHT and other three popular random projection methods. Our results also demonstrate that our proposed method only slightly increase the running time but results in more effective embedding. In the future, we would like to extend our proposed ISRHT methods to nonlinear classification problems. Another interesting direction is to design data-dependent sparse embedding methods. 

%
%

\section{Acknowledgments}
We would like to thank the anonymous reviewers for their insightful comments and valuable suggestions on our paper. This work was supported by NSFC 61906161.

\begin{small}
\bibliographystyle{aaai}
\bibliography{myRef}
\end{small}

\end{document}